%% file: main.tex
\documentclass[10pt,twocolumn,letterpaper]{article}

\usepackage{cvpr}              

\usepackage{makecell} 
\usepackage[table]{xcolor}
\usepackage{times}
\usepackage{latexsym}
\usepackage{multirow}
\usepackage{colortbl}
\usepackage[T1]{fontenc}
\usepackage{wrapfig}
\usepackage{pifont}
\usepackage[utf8]{inputenc}
\usepackage{tabularx}
\usepackage{microtype}

\usepackage{inconsolata}

\usepackage{graphicx}
\usepackage{booktabs}
\usepackage{adjustbox}
\usepackage{amssymb}
\usepackage{microtype}
\usepackage{bbding}
\usepackage{algorithm}
\usepackage{algpseudocode}

\definecolor{cvprblue}{rgb}{0.21,0.49,0.74}
\usepackage[pagebackref,breaklinks,colorlinks,allcolors=cvprblue]{hyperref}

\title{Bridging Visual Representation and
Reinforcement Learning from \\ Verifiable Rewards
in Large Vision-Language Models}

\author{
Yuhang Han$^{1,3,*}$ \quad 
Yuyang Wu$^{1}$ \quad 
Zhengbo Jiao$^{1}$ \quad 
Yiyu Wang$^{1,3}$ \quad 
Xuyang Liu$^{1}$ \\ 
\quad 
Shaobo Wang$^{1}$ \quad 
Hanlin Xu$^{2}$ \quad 
Xuming Hu$^{3}$ \quad 
Linfeng Zhang$^{1,\dagger}$ \\
$^{1}$EPIC Lab, SJTU \quad
$^{2}$Huawei \quad
$^{3}$HKUST (GZ) \\ 
$^{*}$Project leader \quad $^{\dagger}$Corresponding author \\
\textbf{Homepage:} \url{https://kawhiiiileo.github.io/KAWHI_PAGE/  }
}

\begin{document}
\maketitle

\input{sec/0_abstract}
\input{sec/1_introduction}
\input{sec/2_related_work}
\input{sec/3_method}

\input{sec/4_experients}

\input{sec/5_conclutions}


{
    \small
    \bibliographystyle{ieeenat_fullname}
    \bibliography{main}
}
\appendix

\clearpage
\input{sec/6_appendix}


\end{document}

%% file: sec/0_abstract.tex
\begin{abstract}

Reinforcement Learning from Verifiable Rewards (RLVR) has substantially enhanced the reasoning capabilities of large language models in abstract reasoning tasks. However, its application to Large Vision-Language Models (LVLMs) remains constrained by a structural representational bottleneck. Existing approaches generally lack explicit modeling and effective utilization of visual information, preventing visual representations from being tightly coupled with the reinforcement learning optimization process and thereby limiting further improvements in multimodal reasoning performance.
To address this limitation, we propose \texttt{KAWHI} (\textbf{K}ey-Region \textbf{A}ligned \textbf{W}eighted \textbf{H}armonic \textbf{I}ncentive), a plug-and-play reward reweighting mechanism that explicitly incorporates structured visual information into uniform reward policy optimization methods (\textit{e.g.}, GRPO and GSPO). The method adaptively localizes semantically salient regions through hierarchical geometric aggregation, identifies vision-critical attention heads via structured attribution, and performs paragraph-level credit reallocation to align spatial visual evidence with semantically decisive reasoning steps. Extensive empirical evaluations on diverse reasoning benchmarks substantiate \texttt{KAWHI} as a general-purpose enhancement module, consistently improving the performance of various uniform reward optimization methods. Project page: \href{https://kawhiiiileo.github.io/KAWHI_PAGE/}{\texttt{KAWHI}}
\end{abstract}

%% file: sec/1_introduction.tex
\section{Introduction}

\begin{figure}[htbp]
    \centering
    \includegraphics[width=\linewidth]{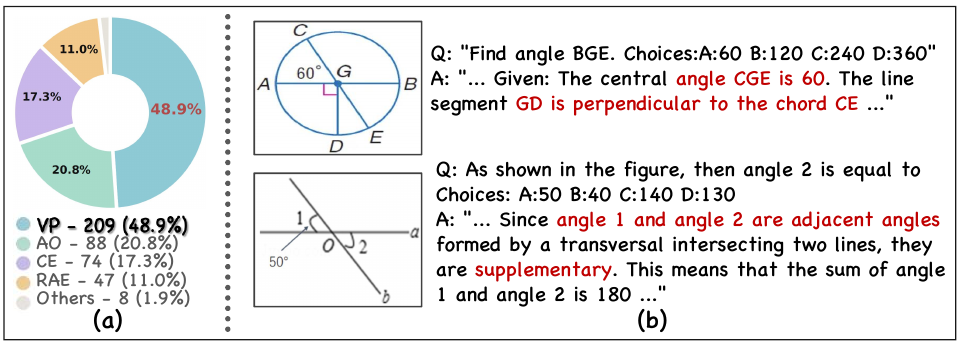}
    \caption{(a) Error distribution across five categories: VP (visual perception error), AO (answer only), CE (calculation error), RAE (rule application error), and Others. (b) Two VP cases from MathVerse. \textcolor{red}{Red} highlights indicate visual misinterpretation.}
    \label{fig:intro}
\end{figure}
Reinforcement Learning from Verifiable Rewards (RLVR), particularly as instantiated through online training paradigms such as Group Relative Policy Optimization (GRPO), has substantially advanced the logical deduction capabilities of Large Language Models (LLMs) within abstract reasoning scenarios confined to unimodal textual contexts~\cite{RLVR1,RLVR2,RLVR3,RLVR4,grpo,gspo,dapo}. Recently, this training paradigm has progressively extended into multimodal cognitive domains, thereby stimulating sustained scholarly efforts toward its cross-modal adaptation and theoretical reconfiguration within Large Vision-Language Models (LVLMs)~\cite{qwen25vl,qwen3vl,innovator-vl}. Existing research predominantly advances along three principal technical trajectories: systematic curation of training corpora~\cite{data1,data3,data5,data4,data2,data6,data7}, principled modeling of reward functions~\cite{reward2,reward3,reward1,reward6,reward5,reward4,reward7}, and structural reformulation of underlying optimization algorithms~\cite{other1,other2}. However, although prior studies have achieved certain performance gains in enhancing the reasoning capabilities of LVLMs, they have \textbf{\textit{failed to establish a principled coupling between visual representations and the RLVR optimization paradigm}}, resulting in suboptimal performance improvements.

Recent LVLMs have exhibited impressive Chain-of-Thought (CoT) reasoning capabilities in complex tasks such as mathematical problem solving and chart interpretation. However, their effectiveness remains constrained by a fundamental representational bottleneck: the implicit assumption that dense tokenization suffices to achieve exhaustive and homogeneous visual encoding. This premise overlooks \textbf{\textit{the intrinsically sparse spatial distribution and task-contingent relevance that characterize visual scenes}}, thereby preventing models from accurately isolating the discriminative visual evidence essential to downstream reasoning. Such a perceptual bottleneck consequently biases attention allocation in RLVR-trained LVLMs, diminishing sensitivity to visually salient cues and ultimately constraining overall performance.

To empirically substantiate this claim, we conduct a dual diagnostic analysis (Fig.~\ref{fig:intro}) on Qwen2.5-VL-7B-Instruct~\cite{qwen25vl} under the GRPO method. At the quantitative level (Fig.~\ref{fig:intro}(a)), we employ Qwen3-VL-30B-A3B~\cite{qwen3vl} as an arbiter to perform fine-grained attribution analysis of erroneous responses on MathVerse~\cite{mathverse}. The results indicate that VP errors constitute 48.9\% of failures, substantially exceeding other categories such as CE and RAE (see Fig.~\ref{fig:intro} caption for definitions). This disparity underscores that \textbf{deficiencies in visual encoding}, rather than peripheral noise factors, represent the primary bottleneck constraining overall model performance. Complementing these quantitative findings, qualitative case analysis (Fig.~\ref{fig:intro}(b)) further reveals the underlying mechanism. Specifically, \textbf{systematic misinterpretation of discriminative visual evidence} (see the red-highlighted region in Fig.~\ref{fig:intro}(b)) precipitates cascading failures along the reasoning chain, substantiating the proposition that misalignment between sparsely distributed visual signals and dense encoding strategies structurally distorts multimodal reasoning dynamics.

Motivated by this diagnosis, recent efforts~\cite{VPPO,credit} have sought to alleviate the bottleneck through fine-grained reward modeling. However, these approaches remain constrained by two fundamental limitations:
\textbf{(i) Semantic Indirection}: VPPO~\cite{VPPO} approximates visual dependency via indirect statistical surrogates (\textit{e.g.}, divergence-based masking), without explicitly grounding reward signals in structured visual semantics.
\textbf{(ii) Architectural Incompatibility}: AT-RL~\cite{credit} requires explicit extraction of cross-modal attention weights, which conflicts with efficiency-oriented operators such as flashattention~\cite{flashattention}, thereby undermining scalable training.

Based on the above analysis, we propose the \textbf{K}ey-Region \textbf{A}ligned \textbf{W}eighted \textbf{H}armonic \textbf{I}ncentive (\texttt{KAWHI}), a plug-and-play, visually-perceived reward reweighting mechanism integrable into uniform reward policy optimization methods~\cite{grpo,gspo,sapo,dapo}. \texttt{KAWHI} adaptively localizes semantically salient regions through hierarchical geometric aggregation and local structural modeling, thereby establishing spatial priors oriented toward visual representation to effectively mitigate information sparsity and foreground-background distributional imbalance. Furthermore, based on MME~\cite{MME} evaluation, we identify vision-correlated attention heads as anchors. The mechanism then leverages image-critical states as semantic identifiers and response query states as information probes to construct passage-level weighting metrics, enabling fine-grained credit assignment across reasoning steps. In summary, our contributions are threefold:

\begin{enumerate}
    \item \textbf{Structural Bottleneck Identification:} We systematically identify critical bottlenecks in applying RLVR to LVLMs and, for the first time, explicitly incorporate visual-geometric information into reward modeling, substantially enhancing multimodal grounding capabilities.
    
    \item \textbf{Vision-Aware Reward Reweighting:} We propose \texttt{KAWHI}, a plug-and-play reward reweighting mechanism compatible with uniform reward policy optimization methods, enabling fine-grained, vision-aware credit assignment without architectural modifications.
    
    \item \textbf{Consistent Performance Gains:} Through extensive evaluations on mathematical and chart-based benchmarks, we demonstrate significant improvements in reasoning accuracy while mitigating hallucination, validating the effectiveness and generalizability of our approach.
\end{enumerate}

%% file: sec/2_related_work.tex
\section{Related Work}

\noindent\textbf{Visual Token Selection.} Visual token selection~\cite{Ficoco,liu2026globalcom2,liu2025mixkv,liu2025vidcom2,alvar2025divprune,zhang2025sparsevlm,chen2025v2drop} research offers valuable insights into identifying critical visual regions. ToMe~\cite{bolya2022token} merges similar tokens to accelerate ViT inference. FastV~\cite{chen2024image} provides a training-free framework for visual token pruning. Through one-to-many token aggregation, FiCoCo~\cite{Ficoco} develops distinct compression methods for visual and linguistic representations. ~\cite{wen2025token} critically examine token pruning approaches in multimodal large language models. DivPrune~\cite{alvar2025divprune} introduces diversity-aware pruning for vision-language models. SparseVLM~\cite{zhang2025sparsevlm} focuses on visual token sparsification for efficient inference. VFlowOpt~\cite{yang2025vflowopt} guides token pruning via visual information flow. These studies collectively indicate that not all visual tokens contribute equally, and identifying high-information-density regions is key to efficiency. However, current strategies are primarily adapted for the texture-dense characteristics of natural images and are mainly employed for \textbf{inference acceleration} in LVLMs, without accounting for the pixel-level sparsity and geometric structural priors inherent in structured visual scenarios (e.g., charts and geometric diagrams).

\noindent\textbf{Visual-Text Alignment.} Visual-text alignment research reveals intrinsic mechanisms within multimodal reasoning processes. RefCOCO~\cite{yu2016modeling} establishes benchmarks for modeling context in referring expressions. Attention Rollout~\cite{abnar2020quantifying} quantifies attention flow in transformers. GLIP~\cite{li2022grounded} unifies grounded language-image pre-training. GroundingDINO~\cite{liu2024grounding} marries DINO with grounded pre-training for open-set detection. SparseMM~\cite{SparseMM} analyzes sparse attention heads in multimodal language models. Structured attention for document understanding~\cite{liu2025structured} highlights pattern regularity in document-level tasks. VISTA~\cite{li2025vista} enhances alignment via mutual information maximization. Attention Re-Alignment~\cite{chen2025attention} addresses suppressed visual grounding through intermediate layer guidance. Spatial-MLLM~\cite{Spatial-MLLM} designs frameworks specifically for spatial reasoning. Spatial Preference Rewarding~\cite{qiu2025spatial} constructs reward functions for spatial understanding tasks. These findings highlight that disconnection between text and image is a core bottleneck constraining multimodal reasoning performance. However, existing alignment insights are mostly utilized for post-hoc interpretation or static supervision in SFT, lacking research on \textbf{integrating visual alignment quality as a dynamic signal} into the optimization process.

\noindent\textbf{RL Optimization Methods.} RL optimization methods have increasingly focused on improving credit assignment efficiency in complex reasoning tasks. PPO~\cite{ppo} establishes a stable foundation for proximal policy optimization through clipped surrogate objectives. DPO~\cite{dpo} simplifies the alignment pipeline by directly optimizing preferences without explicit reward modeling. GRPO~\cite{grpo} removes the critic network and performs group-relative comparisons to enhance training efficiency. VinePPO~\cite{vineppo} introduces fine-grained credit assignment with value networks to better capture token-level contributions. DAPO~\cite{dapo} improves stability via decoupled clipping and dynamic sampling, while GSPO~\cite{gspo} further stabilizes updates through group-level clipping mechanisms. SAPO~\cite{sapo} proposes a soft adaptive policy optimization strategy tailored for language models. Beyond the 80/20 Rule~\cite{8020} emphasizes the importance of high-entropy minority tokens in driving effective learning. VPPO~\cite{VPPO} reweights advantages through visually grounded signals, and Step-wise GRPO~\cite{stepgrpo} refines advantage estimation at the step level. AT-RL~\cite{credit} leverages cross-modality connectivity to enable more precise reinforcement learning. Overall, these studies indicate that, compared with uniform signal allocation strategies, fine-grained signal modulation can more effectively enhance the quality and stability of policy updates. However, most existing approaches do not explicitly model or fully exploit visual modality information, thereby limiting their potential in multimodal reasoning scenarios. Moreover, they generally lack an \textbf{explicit spatial credit assignment mechanism} that incorporates geometric spatial priors, making it difficult to capture higher-level structured dependencies.

%% file: sec/3_method.tex
\section{Method}

\begin{figure*}[t]
    \centering
    \includegraphics[width=\linewidth]{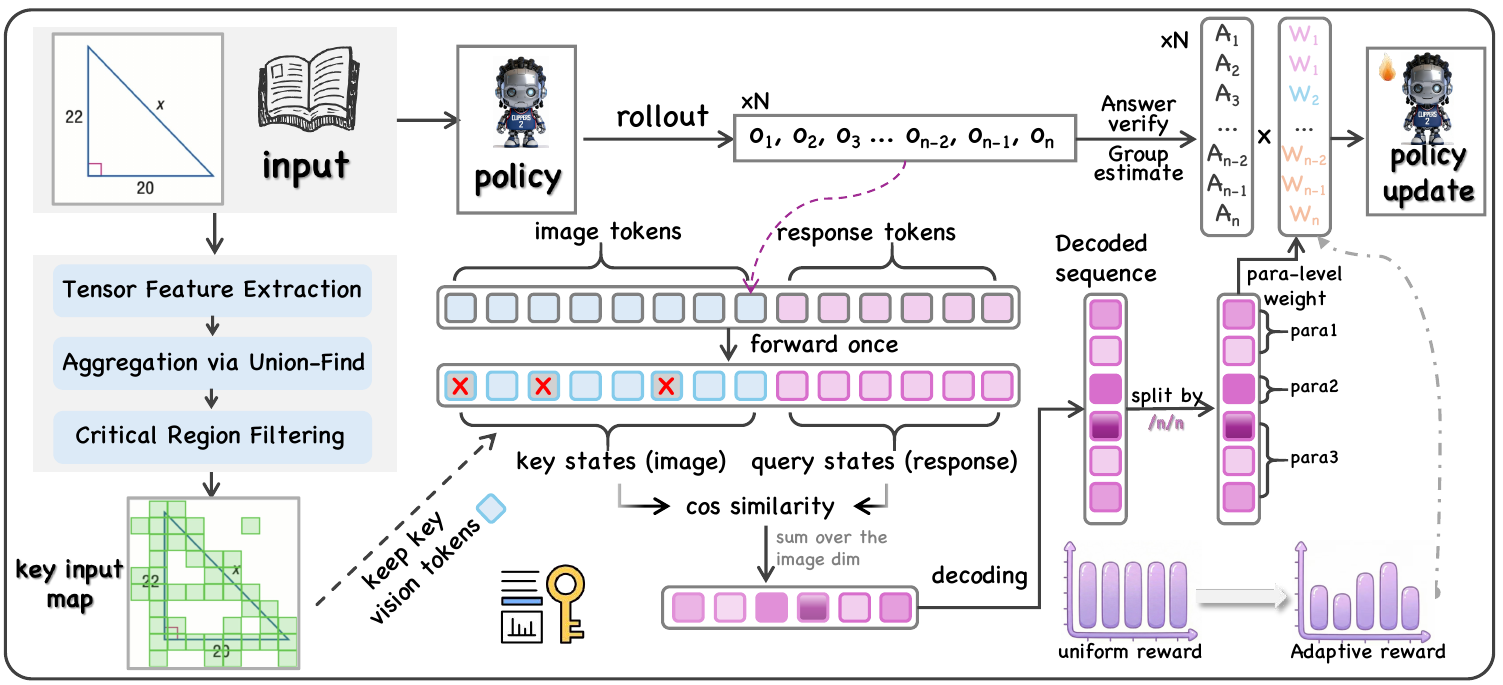}
    \caption{Overview of the \texttt{KAWHI} mechanism. Critical regions are selected using the SGUF algorithm (see section~\ref{sec:SGUF} for details). The decoded sequence denotes the textual output generated from response tokens, segmented by the paragraph delimiter \texttt{\textbackslash n\textbackslash n}.}
    \vspace{-6mm}
    \label{fig:method}
\end{figure*}

In this paper, we propose \texttt{KAWHI} (Fig.~\ref{fig:method}). This section details its core components. We first review GRPO~\cite{grpo} and the uniform reward mechanism (Sec.~\ref{sec:pre}), then introduce the visual token selection algorithm (Sec.~\ref{sec:SGUF}) and the evaluation metrics for critical visual tokens (Sec.~\ref{sec:metric}). Finally, we describe how these metrics are integrated into the reasoning process (Sec.~\ref{sec:weight}).
\subsection{Preliminary: Group-Relative Policy Optimization (GRPO)}
\label{sec:pre}

Given a multimodal prompt $x=(I,q)$, GRPO performs critic-free policy optimization by constructing a baseline from a within-prompt rollout pool.
Let $\pi_{\theta_{\mathrm{old}}}$ sample $G$ completions $\{y_g\}_{g=1}^G$, where
$y_g=(y_{g,1},\ldots,y_{g,T_g})$ and $T_g$ is its length.
A verifier assigns a sequence-level reward $R_g\in\mathbb{R}$ (typically $R_g\in\{0,1\}$).

Rewards are standardized within the group to obtain a relative advantage:
\begin{equation}
\label{eq:grpo-adv}
\mu=\frac{1}{G}\sum_{j=1}^{G}R_j,\quad
\sigma=\sqrt{\frac{1}{G}\sum_{j=1}^{G}(R_j-\mu)^2},\quad
A_g=\frac{R_g-\mu}{\sigma+\epsilon},
\end{equation}
where $\epsilon>0$ ensures numerical stability.
Since rewards are sequence-level, the advantage is broadcast across timesteps:
$A_{g,t}=A_g$.

The token-wise importance ratio is
\begin{equation}
\label{eq:grpo-ratio}
r_{g,t}(\theta)=
\frac{\pi_{\theta}(y_{g,t}\mid x,y_{g,<t})}
{\pi_{\theta_{\mathrm{old}}}(y_{g,t}\mid x,y_{g,<t})}.
\end{equation}

GRPO maximizes a PPO-style clipped surrogate:
\begin{equation}
\label{eq:grpo-token-loss}
\ell_{g,t}(\theta)=
\min\!\Big(
r_{g,t}(\theta)A_{g,t},\
\mathrm{clip}(r_{g,t}(\theta),1-\varepsilon,1+\varepsilon)\,A_{g,t}
\Big),
\end{equation}

and optimizes
\begin{equation}
\label{eq:grpo-objective}
J_{\mathrm{GRPO}}(\theta)=
\mathbb{E}_{x,\{y_g\}\sim\pi_{\theta_{\mathrm{old}}}}
\!\left[
\frac{1}{G}\sum_{g=1}^{G}
\frac{1}{T_g}\sum_{t}\ell_{g,t}(\theta)
\right],
\end{equation}
where $\varepsilon$ is the clipping parameter.

\subsection{Structure-Guided Union-Find (SGUF)}
\label{sec:SGUF}
\textit{What distinguishes structured visual inputs from natural imagery?}
Unlike natural images characterized by rich textures and high semantic density, many vision-language reasoning tasks involve \textbf{structured visual content} such as mathematical expressions, geometric diagrams, and analytical charts.
These inputs exhibit pronounced \textbf{spatial sparsity} and \textbf{foreground--background imbalance}: informative signals are concentrated in thin strokes, symbols, axes, or curve regions, while large portions of the image remain semantically redundant. Such structural properties introduce substantial redundancy into patch token sequences produced by vision encoders. As a result, computational resources are allocated to visually inactive regions, diluting optimization signals and hindering accurate estimation of policy gradients or advantages.

To address this, we propose \textbf{SGUF}, a geometry-aware token filtering mechanism that adaptively identifies informative visual regions.
By modeling local structural patterns, SGUF provides \textbf{spatial priors} for downstream RL optimization, enabling criticality metrics to concentrate on semantically meaningful regions rather than redundant background areas.

\textbf{Structure Tensor Feature Extraction.}
Given an input image $\mathcal{I}$, we first convert it to the CIELab~\cite{cielab} perceptually uniform color space and extract the luminance channel $L \in [0, 100]$.
After applying Gaussian smoothing to $L$, we compute the gradient field $\nabla L = (\partial_x L, \partial_y L)^\top$ using Sobel operators.
For each image patch $\mathcal{P}_{i,j}$, we characterize its local geometry through the second-order structure tensor, defined as the accumulation of gradient outer products over the patch domain:
\begin{equation}
\mathbf{S}_{i,j} = \sum_{(x,y)\in \mathcal{P}_{i,j}} (\nabla L)(\nabla L)^{\top} =
\begin{pmatrix}
\sum g_x^2 & \sum g_x g_y \\
\sum g_x g_y & \sum g_y^2
\end{pmatrix}.
\end{equation}
In practice, we normalize the tensor entries by the patch area $|\mathcal{P}_{i,j}|$ to reduce sensitivity to patch size:
\begin{equation}
\mathbf{S}_{i,j} \leftarrow \frac{1}{|\mathcal{P}_{i,j}|}\mathbf{S}_{i,j}.
\end{equation}
Performing eigendecomposition $\mathbf{S}_{i,j} = \mathbf{R}^\top\boldsymbol{\Lambda}\mathbf{R}$ yields eigenvalues $\lambda_{\max}^{i,j} \geq \lambda_{\min}^{i,j} \geq 0$, which provide rotation-invariant descriptors of local structure. Specifically, the condition $\lambda_{\max} \gg \lambda_{\min}$ indicates anisotropic regions (e.g., stroke edges), whereas $\lambda_{\max} \approx \lambda_{\min} \approx 0$ corresponds to flat background regions. Crucially, these eigenvalues remain invariant under rotation and illumination variations, ensuring robustness against document skew and shadow artifacts.

\textbf{Hierarchical Aggregation via Union-Find.}
We cast region aggregation as graph partitioning over the patch grid.
Each $\mathcal{P}_{i,j}$ is a node in $\mathcal{G}=(\mathcal{V},\mathcal{E})$ under 4-connectivity.
We define a heterogeneous similarity metric combining structural and luminance coherence:
\begin{equation}
d(\mathcal{P}_i, \mathcal{P}_j) =
\underbrace{\frac{\|\boldsymbol{\Lambda}_i - \boldsymbol{\Lambda}_j\|_F}
{\max(\lambda_{\max}^i, \lambda_{\max}^j, \epsilon)}}_{\text{structure}}
+ \alpha \cdot
\underbrace{\frac{|L_i - L_j|}{100}}_{\text{luminance}},
\end{equation}
where $\boldsymbol{\Lambda}=\text{diag}(\lambda_{\max},\lambda_{\min})$,
$\|\cdot\|_F$ is the Frobenius norm, and $\alpha$ controls the trade-off.
While $d(\cdot,\cdot)$ provides a unified dissimilarity score, our implementation uses a \emph{dual-threshold} merge rule to ensure conservative aggregation:
two nodes are merged into component $\mathcal{C}_k$ via Union-Find if
\begin{equation}
\begin{aligned}
\frac{\|\boldsymbol{\Lambda}_i - \boldsymbol{\Lambda}_j\|_F}
{\max(\lambda_{\max}^i, \lambda_{\max}^j, \lambda_{\min}^i, \lambda_{\min}^j, \epsilon)}
&< \delta_s \\
\text{and}\quad |L_i-L_j| &< \delta_l,
\end{aligned}
\end{equation}
where $\delta_s$ and $\delta_l$ are the structure and luminance thresholds, respectively.
This enforces transitive aggregation of structurally coherent stroke regions
while avoiding merges between chromatically similar but geometrically distinct areas
(e.g., white backgrounds vs.\ white formula interiors).

\textbf{Key Region Identification and Adaptive Sampling.}
Given connected components $\{\mathcal{C}_k\}$, we measure structural saliency by the average trace of their structure tensors:
\begin{equation}
E(\mathcal{C}_k) =
\frac{1}{|\mathcal{C}_k|}\sum_{\mathcal{P}\in \mathcal{C}_k} \text{tr}(\mathbf{S})
=
\frac{1}{|\mathcal{C}_k|}\sum_{\mathcal{P}\in \mathcal{C}_k}
(\lambda_{\max}+\lambda_{\min}),
\end{equation}
where $E(\mathcal{C}_k)$ reflects the aggregated gradient magnitude within each region.
We set the energy threshold $\tau$ using a median-based statistic over $\{E(\mathcal{C}_k)\}$:
\begin{equation}
\tau = \beta \cdot \mathrm{median}\big(\{E(\mathcal{C}_k)\}\big),
\end{equation}
where $\beta$ is a scaling factor.
This partitions regions into key regions ($E(\mathcal{C}_k)>\tau$) and background regions ($E(\mathcal{C}_k)\le\tau$).

The selected token set $\mathcal{S}$ is constructed via hybrid sampling: retaining all tokens from key regions and sparsely sampling background regions at rate $r_{\text{skip}}$:
\begin{equation}
\mathcal{S} =
\bigcup_{E(\mathcal{C}_k)>\tau}\mathcal{T}_k
\;\cup\;
\bigcup_{E(\mathcal{C}_k)\le\tau}
\text{Sample}(\mathcal{T}_k,1-r_{\text{skip}}),
\end{equation}
where $\mathcal{T}_k$ denotes tokens corresponding to $\mathcal{C}_k$.
This preserves geometrically informative regions while compressing low-information areas, focusing computation on reasoning-relevant content.

\subsection{Metrics for Spatially-Aware Credit Assignment}
\label{sec:metric}
To integrate the key regions identified by SGUF with the RL rollout process for spatially-aware credit assignment, we revisit the functional differentiation of Query, Key, and Value vectors in the self-attention mechanism~\cite{attention}:
\begin{itemize}
\item \textbf{Query (Q):} Serves as an \textbf{information probe}, representing the information retrieval demands of the current generation token regarding visual content.
\item \textbf{Key (K):} Acts as a \textbf{semantic identifier}, carrying the semantic identity of visual tokens to respond to and match external queries.
\item \textbf{Value (V):} Functions as an \textbf{information payload}, containing the actual representations transmitted downstream after attention weighting.
\end{itemize}
\begin{figure}[tb!]
    \centering
    \includegraphics[width=\linewidth]{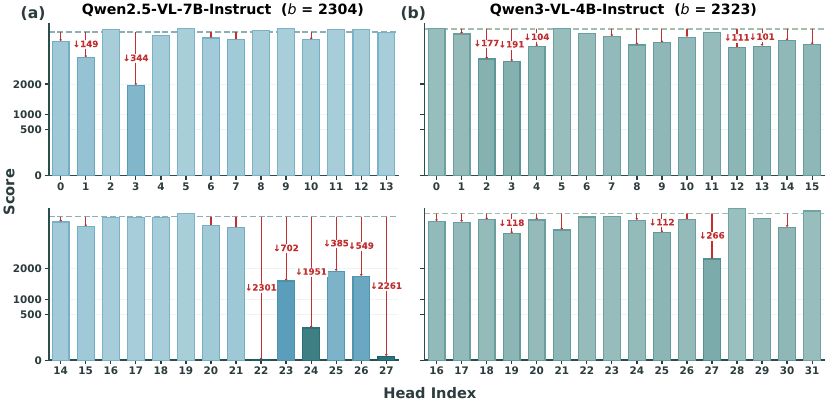}
\caption{Vision-Critical head identification via global ablation on the MME benchmark. Here, b denotes the performance score of the baseline models (Qwen2.5-VL-7B-Instruct\cite{qwen25vl} and Qwen3-VL-4B-Instruct~\cite{qwen3vl}).}
    \label{fig:HeadSelection}
\end{figure}    

Building upon this functional differentiation, we instantiate the generated response tokens from LVLMs as \textbf{Q} (information probes), and the visual tokens within SGUF-selected key regions as \textbf{K} (semantic identifiers). This establishes attentional coupling between generated content and critical visual regions via \textbf{Q-K} matching, thereby quantifying the spatial saliency of the reasoning process.

Nevertheless, the fidelity of this coupling depends on the functional specialization of individual attention heads, which exhibit substantial heterogeneity in their visual contributions~\cite{SparseMM}. To identify the most critical computational units, we perform structured attribution analysis to localize vision-specific heads.

We perform global head ablation on MME~\cite{MME}, which contains 14 subtasks spanning recognition and reasoning (Fig.~\ref{fig:HeadSelection}). Heads with identical indices are masked across all decoder layers to measure their marginal impact on visual performance. Heads whose removal causes an aggregated MME drop exceeding \textbf{100} are defined as \textit{vision-critical}. Using this criterion, we identify attention head subsets in Qwen2.5-VL-7B-Instruct~\cite{qwen25vl} and Qwen3-VL-4B-Instruct~\cite{qwen3vl} that are critical for visual perception.

Leveraging these vision-critical heads, we establish the spatial alignment mechanism between generated responses and SGUF-selected regions. Let $\mathcal{R}$ denote the response token indices and $\mathcal{S} \subseteq \mathcal{I}{\text{select}}$ denote the SGUF-selected key region indices. As query states are not retained in the decoding cache, we conduct an additional forward pass based on the hidden states of the final decoder layer to recover the corresponding query and key states. For Grouped Query Attention~\cite{GQA} with $H_q = g \cdot H_k$ heads, we align the key states with query states by repeating each key head $g$ times. The spatial alignment between response token $t \in \mathcal{R}$ and critical visual token $v \in \mathcal{S}$ is measured by the cosine similarity between $\ell_2$-normalized query and key vectors restricted to $\mathcal{H}{\text{critical}}$:
\begin{equation}
s_{t,v}^{(h)} = \frac{\mathbf{q}_t^{(h)} \cdot \mathbf{k}_v^{(h)}}{\|\mathbf{q}_t^{(h)}\| \|\mathbf{k}_v^{(h)}\|}, \quad h \in \mathcal{H}_{\text{critical}},
\end{equation}
where $\mathbf{q}_t^{(h)} \in \mathbb{R}^d$ is the query vector and $\mathbf{k}_v^{(h)}$ is the corresponding expanded key vector. The aggregated spatial attention score for token $t$ is then computed as:
\begin{equation}
\alpha_t = \frac{1}{|\mathcal{S}| \cdot |\mathcal{H}_{\text{critical}}|} \sum_{h \in \mathcal{H}_{\text{critical}}} \sum_{v \in \mathcal{S}} s_{t,v}^{(h)}.
\end{equation}
This score $\alpha_t \in [-1, 1]$ quantifies the degree to which the generation process focuses on the SGUF-identified critical visual regions through the lens of vision-critical heads, serving as a spatial prior for downstream credit assignment.

\subsection{Semantic-Preserving Spatial Weighting}
\label{sec:weight}

Having established spatially-aware credit assignment metrics, the remaining challenge is to couple these signals effectively with reward allocation over generated responses. The segmentation granularity critically determines the stability and validity of credit attribution. Excessively fine-grained segmentation (e.g., token- or sentence-level) fragments coherent semantic units and disrupts contextual dependencies within the reasoning chain, thereby amplifying noise and destabilizing credit signals. In such cases, rewards become misaligned with semantically coherent reasoning steps and are diffusely assigned to isolated lexical fragments, undermining long-range dependencies and logical consistency. As a result, the spatial saliency metric $\alpha_t$ cannot form a stable correspondence with overall reasoning quality.

To address this issue, we adopt a \textbf{paragraph-level coarse-grained segmentation} strategy. Using paragraph delimiters (\verb|\n\n|) as boundaries, each segment preserves a self-contained reasoning step with intact semantic structure. This design maintains the natural organization of cot reasoning while enabling stable and semantically consistent credit allocation.

Formally, let the response be partitioned into $M$ paragraphs $\{P_j\}_{j=1}^M$, where each paragraph $P_j$ comprises a set of token indices $\mathcal{S}_j$. We first aggregate the spatial saliency scores within each paragraph via mean pooling:
\begin{equation}
\bar{\alpha}_j = \frac{1}{|\mathcal{S}_j|} \sum_{t \in \mathcal{S}_j} \alpha_t.
\end{equation}
Subsequently, these paragraph-level scores are normalized into weights $w_j \in [w_{\min}, w_{\max}]$ through temperature-scaled softmax followed by uniform mixing:
\begin{equation}
\begin{aligned}
\tilde{w}_j &= (1-\lambda) \frac{\exp(\bar{\alpha}_j/\tau)}{\sum_{k=1}^M \exp(\bar{\alpha}_k/\tau)} + \frac{\lambda}{M}, \\
w_j &= w_{\min} + (w_{\max} - w_{\min}) \cdot \tilde{w}_j,
\end{aligned}
\end{equation}
where $\tau$ denotes the temperature parameter controlling the sharpness of the weight distribution, and $\lambda$ is the smoothing coefficient ensuring minimum credit allocation to all paragraphs.

To integrate these spatial priors into the policy optimization objective, we modulate the group-wise baseline advantage by the paragraph-level weights. For a response group $g$ with total reward $R_g$, the baseline advantage is first computed via group-wise standardization:
\begin{equation}
A_g = \frac{R_g - \mu_g}{\sigma_g + \epsilon},
\end{equation}
where $\mu_g$ and $\sigma_g$ denote the mean and standard deviation of rewards within group $g$, and $\epsilon$ is a numerical stability constant. This group-level scalar is then broadcast to all response tokens $t$, yielding the token-level advantage $A_{g,t} = A_g \cdot \mathbb{I}(t \in \mathcal{R}_g)$, where $\mathbb{I}(\cdot)$ is the indicator function for response positions.

Finally, we apply the paragraph-specific weights to modulate the token-level advantages. For token $t$ belonging to paragraph $P_j$ (i.e., $t \in \mathcal{S}_j$), the weighted advantage is computed as:
\begin{equation}
\hat{A}_{g,t} = A_{g,t} \cdot w_j.
\end{equation}
This formulation ensures that tokens within spatially salient paragraphs (high $w_j$) receive amplified gradient signals, while maintaining stable credit allocation across semantically coherent reasoning blocks. The resulting weighted advantages $\hat{A}_{g,t}$ serve as the final surrogate objectives for policy gradient updates, effectively coupling spatial visual attention with linguistic reasoning quality.

%% file: sec/4_experients.tex
\section{Experiments}

\subsection{Experimental Setup}

\subsubsection{Model, training data and evluation}

We conduct experiments on Qwen2.5-VL-7B-Instruct~\cite{qwen25vl} and Qwen3-VL-4B-Instruct~\cite{qwen3vl}. 
For training, Geo3K~\cite{geo3k} is used for mathematical reasoning, while a 20K subset of ChartQA~\cite{chartqa} is constructed for chart understanding (details in Appendix). 
Evaluation is performed on MathVista~\cite{mathvista}, MathVerse~\cite{mathverse}, MathVision~\cite{mathvision}, and WeMath~\cite{wemath} for mathematical reasoning, and on ChartXivDesc~\cite{chartxiv}, ChartXivRea~\cite{chartxiv}, ChartQA~\cite{chartqa}, ChartQA-Pro~\cite{chartqapro}, and ChartMimic~\cite{chartmimic} for chart comprehension. 
Following prior work, we adopt an LLM-as-a-judge protocol with Qwen3-VL-30B-A3B~\cite{qwen3vl} for MathVista, MathVerse, and MathVision, and Exact Match for WeMath. 
All experiments are conducted under a unified setup using the LMMs-Eval~\cite{lmms-eval} framework, and reinforcement learning is performed without any intermediate supervised fine-tuning (SFT).
\subsubsection{Comparison Methods}

We first report the experimental results of fine-grained reward allocation methods, including Step-GRPO~\cite{stepgrpo}, FT-RL~\cite{8020}, and VPPO~\cite{VPPO}. For uniform reward allocation approaches, we present the baseline results of GRPO~\cite{grpo}, DAPO~\cite{dapo}, and GSPO~\cite{gspo}, followed by the performance obtained after incorporating our proposed method. All experiments are conducted on 8 NVIDIA H200 GPUs, with details in the Appendix.
\input{tab/main_results}

\subsection{Main Results}

Table~\ref{main_results} presents comprehensive experiments on Qwen2.5-VL-7B-Instruct~\cite{qwen25vl} and Qwen3-VL-4B-Instruct~\cite{qwen3vl}, integrating \texttt{KAWHI} into the uniform reward-based GRPO~\cite{grpo}, DAPO~\cite{dapo}, and GSPO~\cite{gspo} frameworks. The results show consistent and substantial improvements in average performance across four datasets after incorporating \texttt{KAWHI}. From a methodological perspective, \texttt{KAWHI} explicitly captures critical visual information and adaptively assigns weights across different stages of the reasoning process, thereby enhancing the modeling of key signals in vision–language interactions. This structured attention adjustment reduces irrelevant interference while improving cross-modal alignment, enabling the model to focus more precisely on task-relevant regions during complex reasoning and ultimately enhancing overall perception and reasoning performance.

\input{tab/chart}

Furthermore, we conduct additional validation on chart understanding using Qwen2.5-VL-7B-Instruct with a 20K-sample subset from ChartQA under the GRPO method. As shown in Fig.~\ref{chart_results}, incorporating \texttt{KAWHI} leads to performance gains of \textbf{2.1\%}, \textbf{2.1\%}, \textbf{0.8\%}, \textbf{1.6\%}, and \textbf{2.3\%} on ChartXivDesc~\cite{chartxiv}, ChartXivRea~\cite{chartxiv}, ChartQA~\cite{chartqa}, ChartQA-Pro~\cite{chartqapro}, and ChartMimic~\cite{chartmimic}, respectively. These results indicate that \texttt{KAWHI} generalizes well to structured visual tasks such as chart comprehension and consistently provides stable improvements. 
and improving cross-modal alignment.
\vspace{0mm}

\input{tab/ablation}

\subsection{Ablation Studies}

To evaluate the effectiveness of \texttt{KAWHI}, we conduct five ablation studies (Table~\ref{tab:ablation_mathverse_mathvision}) on Qwen-2.5-VL-7B-Instruct~\cite{qwen25vl}, using GRPO as the baseline and reporting results on MathVerse and MathVision. The ablations cover the following components: \ding{172}~\textit{Region Selection}, \ding{173}~\textit{Reward Metric}, \ding{174}~\textit{Response Granularity}, \ding{175}~\textit{Critical Visual Head}, and \ding{176}~\textit{Positional Encoding}.

\textbf{\textit{\ding{172} Region Selection.}}  
Under the \textit{Region Selection} configuration, we evaluate whether SGUF effectively identifies critical visual tokens that contribute to performance gains. We examine three settings: \textbf{All vision tokens}, \textbf{Random}, and \textbf{Inverse selection}. \textbf{All vision tokens} improves performance over GRPO by 0.64\% on MathVerse and 0.16\% on MathVision, indicating that although retaining complete visual information avoids missing important regions, redundant tokens dilute informative signals. \textbf{Random} achieves gains of 0.4\% and 0.7\%, suggesting that random sampling may occasionally capture critical regions but lacks stability. In contrast, \textbf{Inverse selection} underperforms GRPO on both benchmarks, demonstrating that attending to visually irrelevant regions introduces noise and degrades performance.

\textbf{\textit{\ding{173} Reward Metric.}}  
We compare different similarity computation schemes for reward modeling. Computing similarity between \textit{image (key states)} and \textit{response (key states)} results in performance drops of 0.48\% and 0.33\% relative to \texttt{KAWHI}. Although the \textit{Key--Key} formulation preserves basic evaluation ability, it is less discriminative than the \textit{Key (image)--Query (response)} scheme, as it relies on symmetric static matching. In contrast, \textit{Key--Query} supports asymmetric, task-conditioned alignment, enabling more precise cross-modal modeling.

\textbf{\textit{\ding{174} Response Granularity.}}  
We further investigate finer-grained response partition strategies, including sentence-level and token-level segmentation. Compared to \texttt{KAWHI}, sentence-level segmentation yields performance drops of 0.45\% and 2.64\%, while token-level segmentation results in larger decreases of 1.35\% and 2.64\%. These results suggest that preserving semantic structural integrity is critical, as excessively fine-grained segmentation disrupts contextual coherence and weakens reasoning performance.

\textbf{\textit{\ding{175} Critical Visual Head.}}  
When the metric is computed over all attention heads instead of restricting it to vision-related heads, performance drops by 0.99\% and 0.96\%. This suggests that concentrating on critical visual heads suppresses irrelevant signals and enhances discriminative stability.

\textbf{\textit{\ding{176} Positional Encoding.}}  
Using Key and Query states prior to positional encoding leads to performance decreases of 0.48\% and 0.66\%. This indicates that positional encoding incorporates spatial information into the representations, enabling more accurate visual alignment and more reliable reward computation.

\subsection{A Viable Alternative to SGUF}
\begin{figure}[htbp]
    \centering
    \includegraphics[width=\linewidth]{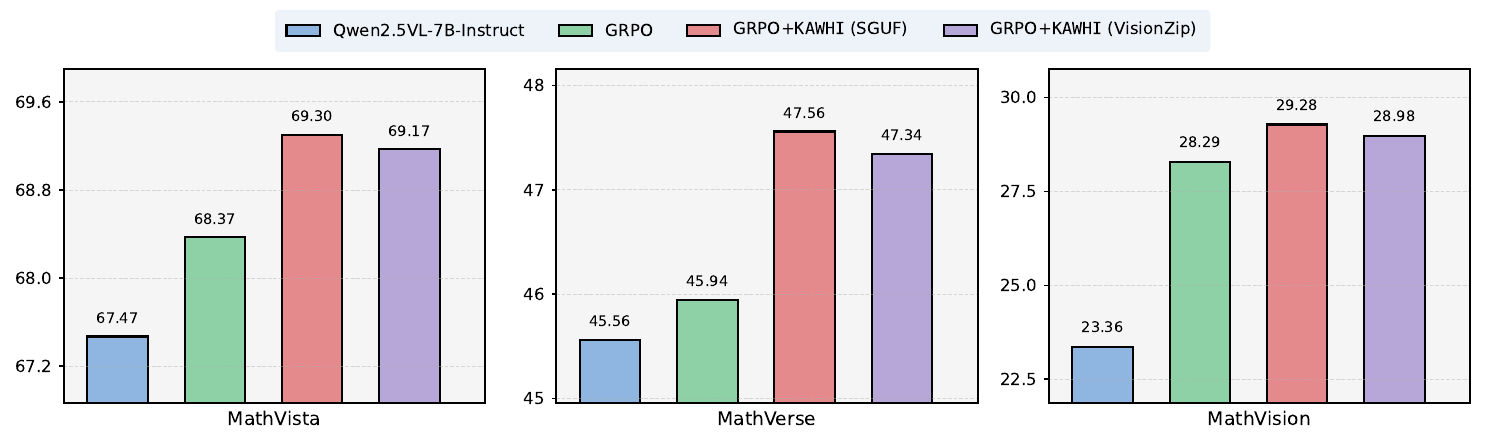}
    \caption{Validation of VisionZip as an effective alternative to SGUF under 50\% visual tokens.}
    \label{sguf_td}
\end{figure}

To examine the transferability of our framework, we replace SGUF with VisionZip~\cite{visionzip} (50\% vision tokens) to obtain salient visual tokens. As shown in Fig.~\ref{sguf_td}, the performance drops on MathVista~\cite{mathvista}, MathVerse~\cite{mathverse}, and MathVision~\cite{mathvision} are only 0.13\%, 0.22\%, and 0.30\%, respectively, indicating that attention-based token selection~\cite{Ficoco,bolya2022token,visionzip} remains compatible with \texttt{KAWHI} and confirming the transferability of \texttt{KAWHI}.

\subsection{Efficiency Analysis}
\begin{figure}[t]
    \centering
    \includegraphics[width=\linewidth]{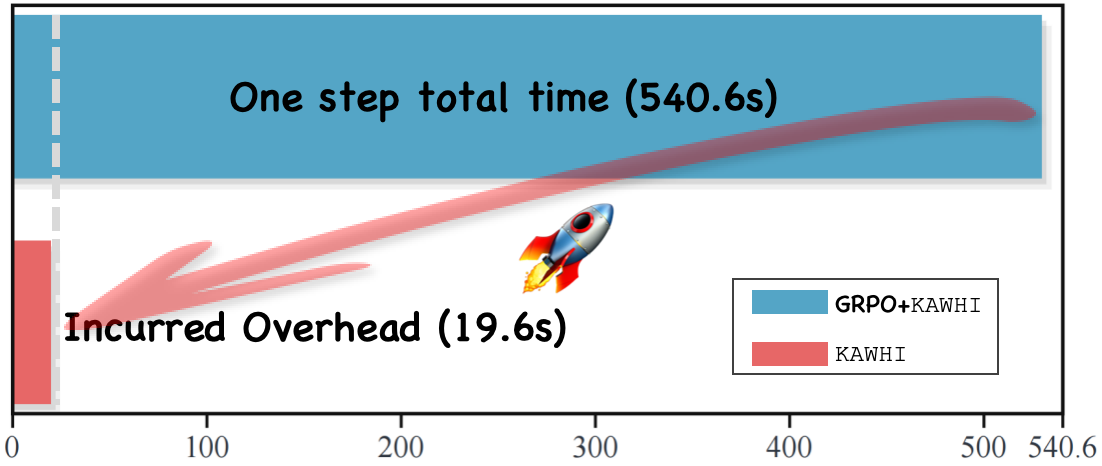}
    \caption{Computational Efficiency Analysis of \texttt{KAWHI}. Component-wise isolation analysis using GRPO to assess the additional computational complexity introduced by our method.}
    \label{effciency_ana}
\end{figure}
To assess the computational efficiency of the proposed method, we conduct rigorous ablation studies measuring per-step training latency under identical hardware configurations and training pipelines (As shown in Fig.~\ref{effciency_ana}). Specifically, within the Qwen2.5-VL-7B-Instruct~\cite{qwen25vl} and VERL~\cite{verl} framework, we compare executions with (R1) and without (R2) \texttt{KAWHI}, yielding per-step durations of 521.0s and 540.6s, respectively. The induced temporal overhead of 19.6s, which is attributable to the integrated weighting mechanism, constitutes merely 3.6\% relative to the baseline, demonstrating that our approach achieves substantial performance gains while incurring only marginal computational costs.

\subsection{Qualitative Experimental Analysis}
\begin{figure}[t]
    \centering
    \includegraphics[width=1\linewidth]{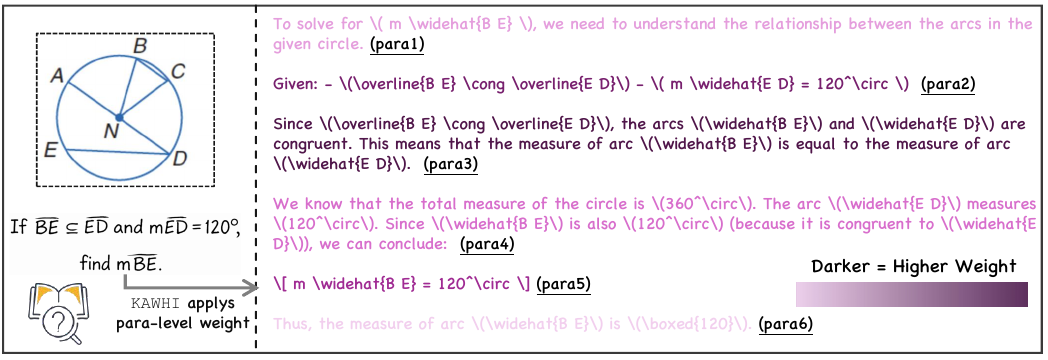}
    \caption{Qualitative Analysis of Weight Allocation in \texttt{KAWHI}. Results are illustrated using cases from Geo3K~\cite{geo3k}. Color intensity represents weight magnitude, where darker colors indicate larger weights, illustrating the allocation behavior of \texttt{KAWHI}.}
    \label{fig:case}
\end{figure}  

Fig.~\ref{fig:case} visualizes the paragraph-level weight distribution induced by \texttt{KAWHI} under GRPO (Qwen2.5-VL-7B-Instruct). Splitting the solution into para1--para6 by \texttt{\textbackslash n\textbackslash n} reveals clear functional roles along the reasoning chain. \textbf{Para2} states the key constraints (congruent chords and the given arc measure), providing the factual basis for inference. \textbf{Para3} applies the decisive rule---congruent chords subtend congruent arcs---to equate the target arc with the known arc. Together, \textbf{para2}--\textbf{para3} constitute constraint injection and rule application. In contrast, \textbf{para1} is introductory and \textbf{para6} restates the answer, contributing minimally to the derivation. \textbf{Para4} mentions the circle’s total measure (360$^\circ$) but is unnecessary and adds limited information. Accordingly, \texttt{KAWHI} assigns substantially higher weights to \textbf{para2} and \textbf{para3} while down-weighting auxiliary or redundant segments. This distribution suggests a positive alignment between paragraph weight and structural importance, supporting the effectiveness of \texttt{KAWHI}.

%% file: tab/main_results.tex
\begin{table*}[t!]
\centering

\captionsetup{font=small,skip=4pt}
\caption{Main results on \textbf{Qwen2.5-VL-7B-Instruct} and \textbf{Qwen3-VL-4B-Instruct}. \textcolor[HTML]{186070}{$+\Delta$}/\textcolor{red}{$-\Delta$} denote changes vs. baselines; \textbf{\textcolor[HTML]{FF6600}{(+X.XX)}} shows gains over base RL methods. Bold = best per model size; gray rows = ours.}

\setlength{\tabcolsep}{10pt}       
\small
\renewcommand{\arraystretch}{1.2}

\setlength{\aboverulesep}{0.4ex}
\setlength{\belowrulesep}{0.4ex}
\setlength{\extrarowheight}{0pt}

\begin{tabular}{@{}lccccc@{}}
\toprule
\textbf{Method} & \textbf{MathVista} & \textbf{MathVerse} & \textbf{MathVision} & \textbf{WeMath} & \textbf{Avg} \\
\midrule
\multicolumn{6}{@{}l}{\textbf{\textit{Qwen2.5-VL-7B-Instruct}}} \\
Base & 67.47 & 45.56 & 23.36 & 54.32 & 47.68 \\
\midrule
\multicolumn{6}{@{}l}{\textit{fine-grained reward}} \\
Step-GRPO & 66.87$_{\text{\tiny\textcolor{red}{$-$0.60}}}$ & 44.32$_{\text{\tiny\textcolor{red}{$-$1.24}}}$ & 24.76$_{\text{\tiny\textcolor[HTML]{186070}{$+$1.40}}}$ & 55.32$_{\text{\tiny\textcolor[HTML]{186070}{$+$1.00}}}$ & 47.82$_{\text{\tiny\textcolor[HTML]{186070}{$+$0.14}}}$ \\
FT-RL & 67.68$_{\text{\tiny\textcolor[HTML]{186070}{$+$0.21}}}$ & 46.88$_{\text{\tiny\textcolor[HTML]{186070}{$+$1.32}}}$ & 26.12$_{\text{\tiny\textcolor[HTML]{186070}{$+$2.76}}}$ & 56.78$_{\text{\tiny\textcolor[HTML]{186070}{$+$2.46}}}$ & 49.37$_{\text{\tiny\textcolor[HTML]{186070}{$+$1.69}}}$ \\
VPPO & 68.42$_{\text{\tiny\textcolor[HTML]{186070}{$+$0.95}}}$ & 47.21$_{\text{\tiny\textcolor[HTML]{186070}{$+$1.65}}}$ & 28.42$_{\text{\tiny\textcolor[HTML]{186070}{$+$5.06}}}$ & 57.12$_{\text{\tiny\textcolor[HTML]{186070}{$+$2.80}}}$ & 50.29$_{\text{\tiny\textcolor[HTML]{186070}{$+$2.61}}}$ \\
\midrule
\multicolumn{6}{@{}l}{\textit{uniform reward}} \\
GRPO & 68.37$_{\text{\tiny\textcolor[HTML]{186070}{$+$0.90}}}$ & 45.94$_{\text{\tiny\textcolor[HTML]{186070}{$+$0.38}}}$ & 28.29$_{\text{\tiny\textcolor[HTML]{186070}{$+$4.93}}}$ & 58.10$_{\text{\tiny\textcolor[HTML]{186070}{$+$3.78}}}$ & 50.18$_{\text{\tiny\textcolor[HTML]{186070}{$+$2.50}}}$ \\
\rowcolor{gray!15} \quad +\texttt{KAWHI} (Ours) & 69.30$_{\text{\tiny\textcolor[HTML]{186070}{$+$1.83}}}$ & 47.56$_{\text{\tiny\textcolor[HTML]{186070}{$+$2.00}}}$ & 29.28$_{\text{\tiny\textcolor[HTML]{186070}{$+$5.92}}}$ & 57.84$_{\text{\tiny\textcolor[HTML]{186070}{$+$3.52}}}$ & 51.00$^{\textbf{\textcolor[HTML]{FF6600}{(+0.82)}}}_{\text{\tiny\textcolor[HTML]{186070}{$+$3.32}}}$ \\
DAPO & 69.20$_{\text{\tiny\textcolor[HTML]{186070}{$+$1.73}}}$ & 47.75$_{\text{\tiny\textcolor[HTML]{186070}{$+$2.19}}}$ & 28.21$_{\text{\tiny\textcolor[HTML]{186070}{$+$4.85}}}$ & 57.98$_{\text{\tiny\textcolor[HTML]{186070}{$+$3.66}}}$ & 50.79$_{\text{\tiny\textcolor[HTML]{186070}{$+$3.11}}}$ \\
\rowcolor{gray!15} \quad +\texttt{KAWHI} (Ours) & 69.08$_{\text{\tiny\textcolor[HTML]{186070}{$+$1.61}}}$ & 48.20$_{\text{\tiny\textcolor[HTML]{186070}{$+$2.64}}}$ & 31.91$_{\text{\tiny\textcolor[HTML]{186070}{$+$8.55}}}$ & 58.21$_{\text{\tiny\textcolor[HTML]{186070}{$+$3.89}}}$ & 51.85$^{\textbf{\textcolor[HTML]{FF6600}{(+1.06)}}}_{\text{\tiny\textcolor[HTML]{186070}{$+$4.17}}}$ \\
GSPO & \textbf{69.42}$_{\text{\tiny\textcolor[HTML]{186070}{$+$1.95}}}$ & 49.87$_{\text{\tiny\textcolor[HTML]{186070}{$+$4.31}}}$ & 27.30$_{\text{\tiny\textcolor[HTML]{186070}{$+$3.94}}}$ & 59.02$_{\text{\tiny\textcolor[HTML]{186070}{$+$4.70}}}$ & 51.40$_{\text{\tiny\textcolor[HTML]{186070}{$+$3.72}}}$ \\
\rowcolor{gray!15} \quad +\texttt{KAWHI} (Ours) & 69.37$_{\text{\tiny\textcolor[HTML]{186070}{$+$1.90}}}$ & \textbf{50.10}$_{\text{\tiny\textcolor[HTML]{186070}{$+$4.54}}}$ & \textbf{32.57}$_{\text{\tiny\textcolor[HTML]{186070}{$+$9.21}}}$ & \textbf{59.71}$_{\text{\tiny\textcolor[HTML]{186070}{$+$5.39}}}$ & \textbf{52.94}$^{\textbf{\textcolor[HTML]{FF6600}{(+1.54)}}}_{\text{\tiny\textcolor[HTML]{186070}{$+$5.26}}}$ \\
\midrule[\heavyrulewidth]
\multicolumn{6}{@{}l}{\textbf{\textit{Qwen3-VL-4B-Instruct}}} \\
Base & 70.43 & 40.86 & 21.38 & 69.43 & 50.53 \\
\midrule
\multicolumn{6}{@{}l}{\textit{uniform reward}} \\
GRPO & 71.07$_{\text{\tiny\textcolor[HTML]{186070}{$+$0.64}}}$ & 41.12$_{\text{\tiny\textcolor[HTML]{186070}{$+$0.26}}}$ & 22.16$_{\text{\tiny\textcolor[HTML]{186070}{$+$0.78}}}$ & 70.31$_{\text{\tiny\textcolor[HTML]{186070}{$+$0.88}}}$ & 51.17$_{\text{\tiny\textcolor[HTML]{186070}{$+$0.64}}}$ \\
\rowcolor{gray!15} \quad +\texttt{KAWHI} (Ours) & 71.80$_{\text{\tiny\textcolor[HTML]{186070}{$+$1.37}}}$ & 42.10$_{\text{\tiny\textcolor[HTML]{186070}{$+$1.24}}}$ & 23.03$_{\text{\tiny\textcolor[HTML]{186070}{$+$1.65}}}$ & 70.98$_{\text{\tiny\textcolor[HTML]{186070}{$+$1.55}}}$ & 51.98$^{\textbf{\textcolor[HTML]{FF6600}{(+0.81)}}}_{\text{\tiny\textcolor[HTML]{186070}{$+$1.45}}}$ \\
DAPO & 69.40$_{\text{\tiny\textcolor{red}{$-$1.03}}}$ & 40.48$_{\text{\tiny\textcolor{red}{$-$0.38}}}$ & 27.63$_{\text{\tiny\textcolor[HTML]{186070}{$+$6.25}}}$ & 69.67$_{\text{\tiny\textcolor[HTML]{186070}{$+$0.24}}}$ & 51.80$_{\text{\tiny\textcolor[HTML]{186070}{$+$1.27}}}$ \\
\rowcolor{gray!15} \quad +\texttt{KAWHI} (Ours) & 71.21$_{\text{\tiny\textcolor[HTML]{186070}{$+$0.78}}}$ & 41.36$_{\text{\tiny\textcolor[HTML]{186070}{$+$0.50}}}$ & 28.43$_{\text{\tiny\textcolor[HTML]{186070}{$+$7.05}}}$ & 71.23$_{\text{\tiny\textcolor[HTML]{186070}{$+$1.80}}}$ & 53.06$^{\textbf{\textcolor[HTML]{FF6600}{(+1.26)}}}_{\text{\tiny\textcolor[HTML]{186070}{$+$2.53}}}$ \\
GSPO & 70.50$_{\text{\tiny\textcolor[HTML]{186070}{$+$0.07}}}$ & 41.62$_{\text{\tiny\textcolor[HTML]{186070}{$+$0.76}}}$ & 32.89$_{\text{\tiny\textcolor[HTML]{186070}{$+$11.51}}}$ & 71.55$_{\text{\tiny\textcolor[HTML]{186070}{$+$2.12}}}$ & 54.14$_{\text{\tiny\textcolor[HTML]{186070}{$+$3.61}}}$ \\
\rowcolor{gray!15} \quad +\texttt{KAWHI} (Ours) & \textbf{72.10}$_{\text{\tiny\textcolor[HTML]{186070}{$+$1.67}}}$ & \textbf{46.82}$_{\text{\tiny\textcolor[HTML]{186070}{$+$5.96}}}$ & 31.09$_{\text{\tiny\textcolor[HTML]{186070}{$+$9.71}}}$ & \textbf{72.15}$_{\text{\tiny\textcolor[HTML]{186070}{$+$2.72}}}$ & \textbf{55.54}$^{\textbf{\textcolor[HTML]{FF6600}{(+1.40)}}}_{\text{\tiny\textcolor[HTML]{186070}{$+$5.01}}}$ \\
\bottomrule
\end{tabular}

\label{main_results}
\vspace{-8pt}
\end{table*}

%% file: tab/chart.tex
\begin{table*}[htbp]
\centering

\captionsetup{font=small,skip=4pt}
\caption{Main results on \textbf{chart understanding benchmarks}. \textcolor[HTML]{186070}{$+\Delta$}/\textcolor{red}{$-\Delta$} denote changes vs. baselines; Bold = best per benchmark; gray rows = ours.}

\setlength{\tabcolsep}{10pt}
\small
\renewcommand{\arraystretch}{1.2}

\setlength{\aboverulesep}{0.4ex}
\setlength{\belowrulesep}{0.4ex}
\setlength{\extrarowheight}{0pt}

\begin{tabular}{@{}lccccc@{}}
\toprule
\textbf{Method} & \textbf{CharXiv Desc} & \textbf{CharXiv Rea} & \textbf{Chart QA} & \textbf{Chart QAPro} & \textbf{Chart Mimic} \\
\midrule
Qwen2.5-VL-7B-Instruct & 67.9 & 42.1 & 87.4 & 41.9 & 40.2 \\
\midrule
\quad + GRPO & 70.1$_{\text{\tiny\textcolor[HTML]{186070}{$+$2.2}}}$ & 44.3$_{\text{\tiny\textcolor[HTML]{186070}{$+$2.2}}}$ & 88.3$_{\text{\tiny\textcolor[HTML]{186070}{$+$0.9}}}$ & 43.7$_{\text{\tiny\textcolor[HTML]{186070}{$+$1.8}}}$ & 42.8$_{\text{\tiny\textcolor[HTML]{186070}{$+$2.6}}}$ \\
\rowcolor{gray!15} \quad + GRPO + \texttt{KAWHI} (Ours) & \textbf{72.2}$_{\text{\tiny\textcolor[HTML]{186070}{$+$4.3}}}$ & \textbf{46.4}$_{\text{\tiny\textcolor[HTML]{186070}{$+$4.3}}}$ & \textbf{89.1}$_{\text{\tiny\textcolor[HTML]{186070}{$+$1.7}}}$ & \textbf{45.3}$_{\text{\tiny\textcolor[HTML]{186070}{$+$3.4}}}$ & \textbf{45.1}$_{\text{\tiny\textcolor[HTML]{186070}{$+$4.9}}}$ \\
\bottomrule
\end{tabular}

\label{chart_results}
\vspace{-8pt}
\end{table*}

%% file: tab/ablation.tex
\begin{table*}[h]
\centering
\caption{\textbf{Ablation on Qwen2.5-VL-7B-Instruct.} Results on MathVerse and MathVision. \textcolor{red}{$\downarrow$X.XX} indicates drop vs. GRPO+~\texttt{KAWHI}.}
\label{tab:ablation_mathverse_mathvision}

\small
\setlength{\tabcolsep}{12pt}
\renewcommand{\arraystretch}{1.15}

\begin{tabular*}{\textwidth}{@{\extracolsep{\fill}}llccc@{}}
\toprule
\textbf{Ablation} & \textbf{Experiment} & \textbf{MathVerse (\%)}$\uparrow$ & \textbf{MathVision (\%)}$\uparrow$ & \textbf{Average} \\
\midrule

-- & GRPO
& 45.94\rlap{$_{\textcolor{red}{\downarrow 1.62}}$}
& 28.29\rlap{$_{\textcolor{red}{\downarrow 0.99}}$}
& 37.11 \\
\rowcolor{gray!15}
-- & GRPO+\texttt{KAWHI} (Ours)
& \textbf{47.56}
& \textbf{29.28}
& \textbf{38.42} \\

\midrule

\multirow{3}{*}{\ding{172} \textit{Region Selection}}
& All vision tokens
& 46.58\rlap{$_{\textcolor{red}{\downarrow 0.98}}$}
& 28.45\rlap{$_{\textcolor{red}{\downarrow 0.83}}$}
& 37.52 \\
& Random
& 46.34\rlap{$_{\textcolor{red}{\downarrow 1.22}}$}
& 28.98\rlap{$_{\textcolor{red}{\downarrow 0.30}}$}
& 37.66 \\
& Inverse selection
& 45.74\rlap{$_{\textcolor{red}{\downarrow 1.82}}$}
& 27.82\rlap{$_{\textcolor{red}{\downarrow 1.46}}$}
& 36.78 \\

\midrule

\ding{173} \textit{Reward Metric}
& Key--Key
& 47.08\rlap{$_{\textcolor{red}{\downarrow 0.48}}$}
& 28.95\rlap{$_{\textcolor{red}{\downarrow 0.33}}$}
& 38.02 \\

\midrule

\multirow{2}{*}{\ding{174} \textit{Response Granularity}}
& Sentence-level
& 47.11\rlap{$_{\textcolor{red}{\downarrow 0.45}}$}
& 26.64\rlap{$_{\textcolor{red}{\downarrow 2.64}}$}
& 36.88 \\
& Token-level
& 46.21\rlap{$_{\textcolor{red}{\downarrow 1.35}}$}
& 26.64\rlap{$_{\textcolor{red}{\downarrow 2.64}}$}
& 36.42 \\

\midrule

\ding{175} \textit{Critical Visual Head}
& w/o Critical Head
& 46.57\rlap{$_{\textcolor{red}{\downarrow 0.99}}$}
& 28.32\rlap{$_{\textcolor{red}{\downarrow 0.96}}$}
& 37.45 \\

\midrule

\ding{176} \textit{Positional Encoding}
& before RoPE
& 47.08\rlap{$_{\textcolor{red}{\downarrow 0.48}}$}
& 28.62\rlap{$_{\textcolor{red}{\downarrow 0.66}}$}
& 37.85 \\

\bottomrule
\end{tabular*}
\end{table*}

%% file: sec/5_conclutions.tex
\section{Conclusions}
This work identifies a structural bottleneck in applying RLVR to LVLMs, arising from the mismatch between sparse, structured visual signals and dense visual token representations. Through quantitative and qualitative analyses, we show that visual perception errors constitute the primary failure mode in multimodal reasoning, indicating that existing reward mechanisms underutilize spatial visual information. To address this limitation, we introduce \texttt{KAWHI}, a plug-and-play reward reweighting module that integrates structured visual region alignment and paragraph-level credit assignment into uniform reward optimization. Experiments on mathematical and chart-based benchmarks demonstrate that \texttt{KAWHI} improves reasoning accuracy while reducing hallucination, highlighting the value of structurally grounded reward modeling in multimodal reinforcement learning.

%% file: sec/6_appendix.tex
\section*{Appendix}  
\addcontentsline{toc}{section}{Appendix}  

\setcounter{subsection}{0}
\renewcommand{\thesubsection}{\Alph{subsection}}
\subsection{Experimental Details}
We implement our method within the VERL~\cite{verl} reinforcement learning framework. All experiments are conducted on 8xH200 GPUs, and Fully Sharded Data Parallel (FSDP) is adopted to improve distributed training efficiency and memory utilization. During reinforcement learning, we employ group-based policy optimization and generate five candidate responses for each input to perform group-wise policy updates.

For the GRPO~\cite{grpo}/GSPO~\cite{gspo} configuration, the learning rate is set to $1\times10^{-6}$, with a training batch size of 512. The maximum prompt length is 1024 tokens and the maximum response length is 2048 tokens. For the DAPO~\cite{dapo} configuration, the overall training pipeline remains the same, but the batch size is increased to 1024 and the maximum response length is limited to 1024 tokens for improved training efficiency.

In our implementation, we enable a key-region analysis module to identify semantically salient visual regions. Specifically, the structural saliency threshold is set to 0.5, the luminance threshold to 30, and the energy threshold to 0.1. Gaussian smoothing ($\sigma = 1.0$) is applied to stabilize the region responses, and a skip ratio of 0.7 is used to filter redundant visual regions. For reasoning supervision, we adopt paragraph-level credit assignment and align visual evidence with textual reasoning via Key--Query cross-modal similarity. The paragraph weights are constrained within the range $[0.1,1.0]$, and cross-modal alignment is computed only on selected attention heads. Specifically, Qwen2.5-VL-7B-Instruct~\cite{qwen25vl} uses $[0,1,3,22\text{--}27]$, while Qwen3-VL-4B-instruct~\cite{qwen3vl} uses $[2,3,4,12,13,19,25,27]$.

\subsection{Hyperparameter Sensitivity Analysis}
\begin{figure}[htbp]
    \centering
    \includegraphics[width=0.5\textwidth]{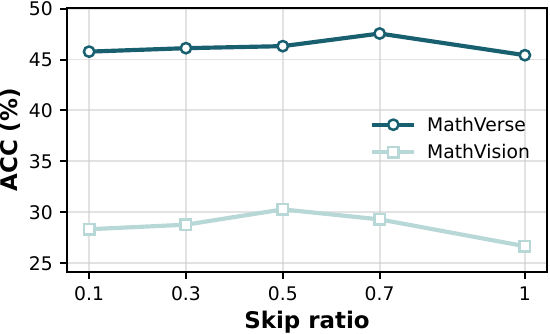}
    \caption{Sensitivity Analysis of the SGUF Hyperparameter \textit{skip\_ratio} (Qwen2.5-VL-7B-Instruct with GRPO+\texttt{KAWHI} on MathVerse and MathVision)}
    \label{hyper_ana}
\end{figure}
After obtaining the set of critical visual tokens via the SGUF algorithm, we conduct a sensitivity analysis on its internal hyperparameter, \textit{skip\_ratio}. This parameter controls the proportion of randomly retained tokens from the unselected visual tokens, thereby balancing the emphasis on critical regions with the preservation of global visual information. Figure 5 presents the performance trends under five different skip ratios. The results indicate that performance peaks when \textit{skip\_ratio} is set to 0.7, achieving an optimal trade-off between key-region enhancement and overall visual information retention. A larger ratio results in too few visual tokens and potential information loss, whereas a smaller ratio introduces excessive redundancy, diluting the contribution of critical regions and weakening the model’s discriminative capability.

\subsection{Theoretical Complexity Analysis}
We analyze the additional computational overhead introduced by our method relative to standard GRPO training. 
Let $H_I \times W_I$ denote the spatial resolution of the input image, $N$ the number of visual tokens produced by the vision encoder, and $N_s \le N$ the number of SGUF-selected key visual tokens. Let $S$ be the full sequence length, $T$ the response length, $M$ the number of paragraphs obtained from response segmentation, $H$ the number of attention heads used for spatial similarity computation, and $d$ the per-head feature dimension. The training cost of standard GRPO is denoted by $\mathcal{F}_{\mathrm{GRPO}}$.

Importantly, the current implementation does not modify the forward pass of the backbone LVLM and therefore does not alter the backbone FLOPs. Instead, the method introduces auxiliary spatial-prior computation on top of standard GRPO. The overall training complexity can thus be written as
\begin{equation}
\mathcal{F}_{\mathrm{ours}} = \mathcal{F}_{\mathrm{GRPO}} + \mathcal{F}_{\mathrm{SGUF}} + \mathcal{F}_{\mathrm{align}} + \mathcal{F}_{\mathrm{para}} .
\end{equation}

\vspace{0.5em}
\noindent\textbf{SGUF Key-Region Extraction.} 
SGUF first performs color-space conversion, Gaussian smoothing, and gradient estimation on the input image, which scales linearly with the image resolution:
\begin{equation}
O(H_I W_I).
\end{equation}
It then performs region aggregation on a patch adjacency graph $G=(V,E)$, where $|V|=N$ and $|E|=O(N)$ under four-neighborhood connectivity. 
Using Union-Find, the region merging complexity is
\begin{equation}
O(|E|\alpha(N)) = O(N\alpha(N)),
\end{equation}
where $\alpha(\cdot)$ denotes the inverse Ackermann function. 
Subsequent region-energy estimation and token sampling incur only linear overhead. 
Therefore,
\begin{equation}
\mathcal{F}_{\mathrm{SGUF}} = O(H_I W_I + N\alpha(N)).
\end{equation}

\vspace{0.5em}
\noindent\textbf{Spatial Alignment.} 
Spatial alignment estimates the relevance between response tokens and SGUF-selected visual regions using final-layer representations. 
Organizing and normalizing the required hidden states contributes
\begin{equation}
O(H S d).
\end{equation}
The dominant computation arises from head-wise similarity evaluation between $T$ response tokens and $N_s$ key visual tokens across $H$ heads:
\begin{equation}
O(H T N_s d).
\end{equation}
Thus,
\begin{equation}
\mathcal{F}_{\mathrm{align}} = O(H S d + H T N_s d).
\end{equation}
When $T$ and $N_s$ are moderately large, the dominant term is $O(H T N_s d)$.

\vspace{0.5em}
\noindent\textbf{Paragraph-Level Reweighting.} 
Paragraph-level reweighting aggregates token-level spatial saliency within each paragraph and broadcasts normalized weights back to the token-level advantages. 
This process involves a linear scan over the response sequence and paragraph-level normalization, yielding
\begin{equation}
\mathcal{F}_{\mathrm{para}} = O(T + M).
\end{equation}

\vspace{0.5em}
\noindent\textbf{Overall Complexity.} 
Combining the above components, the overall complexity becomes
\begin{equation}
\mathcal{F}_{\mathrm{ours}} = \mathcal{F}_{\mathrm{GRPO}} + O(H_I W_I + N\alpha(N) + H S d + H T N_s d + T + M).
\end{equation}
Since $M \ll T$ in practice, both SGUF preprocessing and paragraph reweighting introduce only near-linear overhead, while the dominant additional cost arises from the spatial alignment term $O(H T N_s d)$. Consequently, the proposed method preserves the backbone complexity of GRPO while introducing only modest auxiliary computation for spatially-aware credit assignment.

\subsection{Algorithm Procedure}
Algorithms~\ref{alg:sguf} and~\ref{alg:kawhi} present the pseudocode for our key region selection strategy and the \texttt{KAWHI} training procedure, respectively.

\begin{algorithm}[H]
\caption{SGUF}
\label{alg:sguf}
\begin{algorithmic}[1]
\Require Image $\mathcal{I}$; thresholds $(\delta_s,\delta_l)$; $\sigma,\beta,r_{\text{skip}}$
\Ensure Token set $\mathcal{S}$

\State $L \leftarrow \mathrm{Gaussian}(L,\sigma)$
\State $\mathbf{S}_{i,j}=\sum_{(x,y)\in\mathcal{P}_{i,j}}(\nabla L)(\nabla L)^\top$
\State $\boldsymbol{\Lambda}_{i,j}=\mathrm{eig}(\mathbf{S}_{i,j})$

\State Merge $(i,j)$ if 
$\frac{\|\boldsymbol{\Lambda}_i-\boldsymbol{\Lambda}_j\|_F}
{\max(\lambda_{\max}^i,\lambda_{\max}^j,\epsilon)}<\delta_s$
and $|L_i-L_j|<\delta_l$

\State $\{\mathcal{C}_k\} \leftarrow \mathrm{UnionFind}(\mathcal{P})$

\State $E(\mathcal{C}_k)=\frac{1}{|\mathcal{C}_k|}
\sum_{\mathcal{P}\in\mathcal{C}_k}(\lambda_{\max}+\lambda_{\min})$

\State $\tau=\beta\cdot\mathrm{median}(E(\mathcal{C}_k))$

\State $\begin{aligned}[t]
\mathcal{S} = {} & \bigcup_{E(\mathcal{C}_k)>\tau}\mathcal{T}_k \\
& \cup \bigcup_{E(\mathcal{C}_k)\le\tau}\mathrm{Sample}(\mathcal{T}_k,1-r_{\text{skip}})
\end{aligned}$

\State \Return $\mathcal{S}$
\end{algorithmic}
\end{algorithm}

\begin{algorithm}[H]
\caption{\texttt{KAWHI}}
\label{alg:kawhi}
\begin{algorithmic}[1]
\Require Prompt $x=(I,q)$; rollout size $G$; reward $R(\cdot)$; critical heads $\mathcal{H}_{c}$
\Ensure Updated policy $\theta$

\For{each iteration}
    \State $\mathcal{S}\leftarrow\textsc{SGUF}(I)$
    \State $\{y_g\}_{g=1}^G\sim\pi_{\theta_{\text{old}}}(\cdot|x)$
    \State $R_g\leftarrow R(x,y_g)$
    \State $A_g\leftarrow\frac{R_g-\mu}{\sigma+\epsilon}$

    \ForAll{$y_g$}
        \State $y_g\rightarrow\{P_j\}_{j=1}^M$
        \State $\alpha_t\leftarrow\mathrm{Sim}_{QK}(t,\mathcal{S},\mathcal{H}_c)$
        \State $w_j\leftarrow\mathrm{Norm}(\{\alpha_t:t\in P_j\})$
        \State $\hat{A}_{g,t}\leftarrow A_g\cdot w_j$
    \EndFor

    \State Update $\theta$ via PPO with $\hat{A}_{g,t}$
\EndFor

\State \Return $\theta$
\end{algorithmic}
\end{algorithm}